\documentclass[10pt,conference]{IEEEtran}
\IEEEoverridecommandlockouts
\usepackage{cite}
\usepackage{amsmath,amssymb,amsfonts}
\usepackage{graphicx}
\usepackage{textcomp}
\usepackage{xcolor}

\usepackage{hyperref}

\usepackage{multirow}
\usepackage{adjustbox}
\usepackage{algorithm}
\usepackage{algpseudocode} % loads algorithmicx
\usepackage{booktabs}

\usepackage{tikz}
\usetikzlibrary{arrows.meta, positioning, shapes}
\def\BibTeX{{\rm B\kern-.05em{\sc i\kern-.025em b}\kern-.08em
    T\kern-.1667em\lower.7ex\hbox{E}\kern-.125emX}}
\begin{document}

\IEEEoverridecommandlockouts

\title{Beyond Numerical Features: CNN-Driven Algorithm Selection via Contour Plots for Continuous Black-Box Optimization\\
% \thanks{Accepted by the International Joint Conference on Neural Networks (IJCNN 2026), IEEE World Congress on Computational Intelligence (WCCI 2026). This is the authors' accepted version.}
}

\author{
% \IEEEauthorblockN{Anonymous Author(s)}

\IEEEauthorblockN{
Yiliang Yuan\IEEEauthorrefmark{1},
Xiang Shi\IEEEauthorrefmark{2},
Mustafa Misir\IEEEauthorrefmark{2}
}
\IEEEauthorblockA{\IEEEauthorrefmark{1}
Mohamed bin Zayed University of Artificial Intelligence, Masdar City, United Arab Emirates\\
yiliang.yuan@mbzuai.ac.ae
}
\IEEEauthorblockA{\IEEEauthorrefmark{2}
Duke Kunshan University, Suzhou, China\\
\{xiang.shi, mustafa.misir\}@dukekunshan.edu.cn
}

\thanks{Accepted by the International Joint Conference on Neural Networks (IJCNN 2026), IEEE World Congress on Computational Intelligence (WCCI 2026). This is the authors' accepted version.}
\thanks{\footnotesize © 2026 IEEE. Personal use of this material is permitted. Permission from IEEE must be obtained for all other uses, in any current or future media, including reprinting/republishing this material for advertising or promotional purposes, creating new collective works, for resale or redistribution to servers or lists, or reuse of any copyrighted component of this work in other works.}
}

% \thanks{Accepted by the International Joint Conference on Neural Networks (IJCNN 2026), IEEE World Congress on Computational Intelligence (WCCI 2026). This is the authors' accepted version.}

\maketitle

\begin{abstract}
The present paper introduces a new representation-driven approach to per-instance algorithm selection, applied to black-box optimization, for automatically choosing the most promising solver from a fixed portfolio.
%%%Automated algorithm selection aims to choose, for each black-box optimization instance, the solver in a fixed portfolio that is expected to perform best. 
Prior work in continuous optimization largely relies on numerical descriptors, including Exploratory Landscape Analysis features and learned embeddings such as Deep-ELA. This work studies a complementary representation: contour-map visualizations of probed landscapes. A CNN regressor takes multiple instance-specific contour views (stacked or encoded per view and aggregated) and predicts per-solver performance, enabling selection by the predicted best value. 
% On the standard BBOB 2009 single-objective protocol, the resulting selectors improve substantially over the single best solver and are competitive with strong ELA/DeepELA baselines; Wilcoxon signed-rank tests confirm a significant improvement over SBS. 
On the standard BBOB 2009 single-objective protocol, the resulting selectors significantly outperform the single best solver (SBS) and are competitive with feature-based baselines. %ELA/DeepELA baselines. 
A subsequent bi-objective evaluation under the DeepELA setting further indicates that the same image-based principle can be competitive when using windowed contour views. 
Overall, the results suggest that simple vision models can exploit spatial structure in probed landscapes for algorithm selection without handcrafted ELA features.
\end{abstract}


% © 2026 IEEE. Personal use of this material is permitted. Permission from IEEE must be obtained for all other uses, in any current or future media, including reprinting/republishing this material for advertising or promotional purposes, creating new collective works, for resale or redistribution to servers or lists, or reuse of any copyrighted component of this work in other works.

\section{Introduction}
%%%%%%%%%%%%%%%%%%%%%%%%%%%%%%%%%%%%%%%%%%
%%%%%%%%%%%%%%%%%%%%%%%%%%%%%%%%%%%%%%%%%%
Continuous black-box optimization (BBO) is a common setting in science and engineering where only function evaluations are available and gradient information is inaccessible~\cite{long2023bbob}. 
In such regimes, derivative-free solvers like CMA-ES variants offer strong performance, yet the No Free Lunch theorem implies that no single algorithm universally outperforms across all instances~\cite{585893}. 
% Consequently, performance can often be improved by selecting a solver conditional on the problem instance rather than committing to a static choice, motivating Automated Algorithm Selection (AAS)~\cite{kerschke2019automated,prager2022automated}.  
Consequently, the potential to improve performance by selecting a solver tailored to each problem instance rather than committing to a static choice, motivates Automated Algorithm Selection (AAS)~\cite{kerschke2019automated,prager2022automated}.

AAS in continuous BBO typically relies on instance representations derived from probing evaluations. 
Exploratory Landscape Analysis (ELA) constructs numerical feature vectors intended to summarize properties such as curvature, separability, and modality~\cite{kerschke2019automated1}. 
Recent works replace handcrafted ELA features with learned embeddings, reducing feature correlation and improving robustness under the same benchmark protocol~\cite{seiler2025deep}. 
% However, both lines of work essentially regress from abstract numerical descriptors, which may not preserve spatial structure that is evident when a landscape is visualized.
However, both types of studies rely on abstract numerical descriptors that may fail to capture the spatial structure evident in a visualized landscape.
In parallel, CNN-based AAS has been explored in discrete domains where direct instance encodings are available \cite{loreggia2016deep}; the continuous BBO setting differs in that a representation must be constructed through probing rather than read off from an explicit input.

This paper studies a probing-based, image-centered representation for continuous BBO.
% We mainly want to investigate whether 2-D contour maps contain useful AAS signal in BBO problems.
This paper mainly investigates whether 2-D contour maps of probed landscapes contain useful signal for AAS in continuous BBO.
Each problem instance is rendered as a grayscale contour map from objective evaluations on a fixed grid.
Then, a CNN model is built to predict per-solver performance and to choose the best algorithm.
Specifically, two variants are considered: a combined-view model that stacks multiple instance views as channels, and a separate-view model that encodes each view and aggregates the resulting features. 
Evaluation on BBOB~\cite{bbob2019} under the established protocol shows that our methods substantially improve over the single best solver and are competitive with ELA/Deep-ELA~\cite{seiler2025deep} baselines. 
A follow-up bi-objective study following~\cite{seiler2025deep} further examines the same principle. 

The main contributions are:
\begin{enumerate}
    \item a probing-based AAS formulation for continuous BBO that uses contour-map renderings as instance representations and trains CNN selectors without handcrafted landscape features
    \item an empirical study on single- and multi-objective optimization problems under standard AAS protocols, analyzing the effects of view aggregation and input resolution, and positioning contour-based selection relative to ELA and Deep-ELA baselines. 
\end{enumerate}

The remainder of this article is structured as follows. 
Section~\ref{sec:relatedworks} discusses optimization and AAS. 
Our proposed CNN-based AAS setup and the representation it utilizes are detailed in Section~\ref{sec:Methods}. 
Section~\ref{sec:Experiments} reports the empirical results. 
Finally, Section~\ref{sec:Conclusion} summarizes the work, presents the key findings, and outlines future research directions.

\definecolor{darkgray}{RGB}{102,102,102}
\definecolor{lightgray}{RGB}{221,221,221}

\tikzset{
    block/.style={rectangle, draw=darkgray, fill=lightgray, text width=4.15cm, text centered, rounded corners, minimum height=1.5cm},
    line/.style={draw, -Latex},
    %%cloud/.style={draw=none, ellipse, minimum height=2em},
    cloud/.style={draw=none, rectangle, minimum height=4em},
    outpbox/.style={draw, dashed, rectangle, minimum height=2em},
}

\begin{figure*}[t]
   \centering
   
\begin{tikzpicture}[node distance=1cm, auto]
    
    % Nodes
    \node [block] (problem) {\textbf{Problem Instance Set} $\mathbf{\mathcal{I}} = \{i_1, i_2, \dots, i_n\}$};
    % \node [block, right=of problem] (features) {\textbf{Feature Extractor} \\$\displaystyle \mathbf{I} \rightarrow \mathbf{F}$};
    \node [block, right=of problem] (features) {\textbf{Feature Extractor} \\$\mathbf{\mathcal{I}} \rightarrow \mathbf{Z}$};
    \node [block, below=of features] (model) {\textbf{AAS Model} \\$\displaystyle \Phi: \mathbf{Z} \mapsto \hat{\mathcal{P}}(\mathcal{A}, \mathcal{I})$};
    \node [block, left=of model] (algorithms) {\textbf{Algorithm Portfolio} $\mathcal{A} = \{a_1, a_2, \dots, a_m\}$};
    \node [cloud, align=center, right=of features, xshift=-0.45cm] (new) {a new problem instance \\  $i_{\text{new}}$};
    \node [outpbox, right=of model] (output) {$\Phi: \mathbf{Z}(i_{\text{new}}) \xrightarrow{} a^*$};
    
    % Arrows
    \path [line, line width=1pt] (problem) -- (features);
    \path [line, line width=1pt] (features) -- (model);
    \path [line, line width=1pt] (algorithms) -- (model);
    \path [line, dashed, line width=1pt] (new) -- (output);
    %%\path [line, blue!50] (new) |- (model);
    \path [line, line width=1pt] (model) -- (output);
    
    % % Annotations
    % \node [above=0.2cm of features.north west, anchor=south west, font=\small] {\text{\hspace{0.9cm}Feature representation}};
    % \node [below=0.2cm of model.south west, anchor=north west, font=\small] {\text{Algorithm performance prediction}};
    
\end{tikzpicture}
   \caption{A traditional performance prediction based Automated Algorithm Selection (AAS) process. $\mathcal{I}$ denotes the instance set, $\mathcal{A}$ the algorithm portfolio, and \textbf{Z} the feature extractor, yielding a predicted solver \(a^*\).}
   \label{AS_process}
\end{figure*}
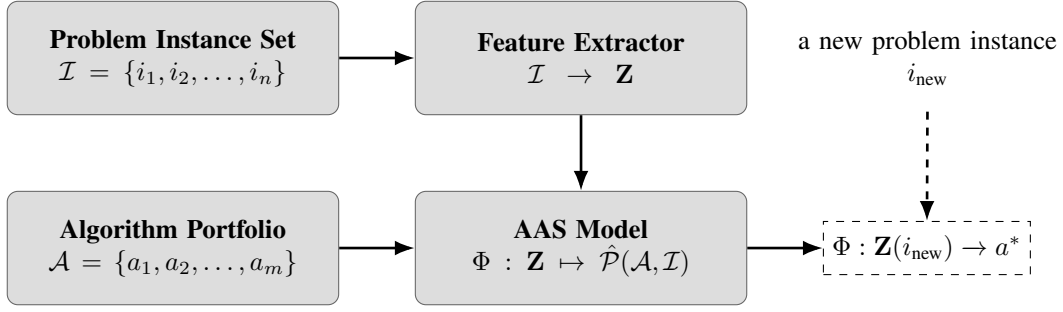

%%%%%%%%%%%%%%%%%%%%%%%%%%%%%%%%%%%%%%%%%%
%%%%%%%%%%%%%%%%%%%%%%%%%%%%%%%%%%%%%%%%%%
\section{Background}  
\label{sec:relatedworks}
%%%%%%%%%%%%%%%%%%%%%%%%%%%%%%%%%%%%%%%%%%  
%%%%%%%%%%%%%%%%%%%%%%%%%%%%%%%%%%%%%%%%%%
\subsection{Optimization Problems}
% An optimization problem involves identifying variable values that maximize or minimize an objective function while adhering to specified constraints~\cite{daskalakis2021complexity}.
% Optimization problems are identified as discrete or continuous based on variables' value characteristics, and as single-objective or multi-objective depending on the number of optimization objectives.
% Discrete optimization problems involve variables that take integer, binary, or categorical values, resulting in combinatorial solution spaces. These problems are characterized by potentially vast yet distinct solution spaces. Classic examples include the Traveling Salesman Problem (TSP), knapsack problem, and scheduling problems.
% The continuous optimization problems deal with variables defined over real numbers, optimizing functions with smooth or non-smooth landscapes.
% Problems often involve minimizing or maximizing objectives like cost, error, or performance metrics. 
% The single-objective optimization targets finding an optimal solution based on the given objective function.
% Multi-objective optimization, on the other hand, addresses conflicting objectives, such as maximizing performance while minimizing cost, with solutions evaluated on a trade-off curve known as the Pareto front, where no objective can be improved without worsening another.

An optimization problem requires determining values for decision variables that minimize or maximize an objective function, possibly subject to constraints~\cite{daskalakis2021complexity}. 
Problems are commonly categorized by variable domain (discrete vs. continuous) and by the number of objectives (single- vs. multi-objective). 
Discrete optimization involves integer, binary, or categorical variables and typically yields combinatorial search spaces. 
Continuous optimization, in contrast, deals with real-valued variables and induces continuous objective landscapes. 
Single-objective optimization targets one scalar objective, whereas multi-objective optimization considers multiple, often conflicting, objectives and evaluates solutions via Pareto dominance and the Pareto front.

%%%%%%%%%%%%%%%%%%%%%%%%%%%%%%%%%%%%%%%%%%%%%%
\subsection{Automated Algorithm Selection for Optimization}
%%%%%%%%%%%%%%%%%%%%%%%%%%%%%%%%%%%%%%%%%%%%%%
% Automated Algorithm Selection (AAS) is a technique for automatically selecting the most suitable algorithm from a set of algorithms for a given problem instance. The foundational principle of AAS is the \textit{No Free Lunch Theorem}, which posits that no single algorithm universally outperforms others across all problem instances~\cite{585893}. In practice, AAS identifies the best algorithm for each instance by extracting features that describe the structural or behavioral properties of optimization problems, and pairing these features with performance data such as runtime or solution quality from various algorithms~\cite{alissa2023automated}. 

% Machine learning models are then trained to map these features to the best-performing algorithm, enabling effective selection for new problem instances. A typical AAS process is depicted in Figure~\ref{AS_process}, where $\mathcal{I}$ represents the set of problem instances, $\mathcal{A}$ denotes the set of candidate algorithms, and \textbf{F} is the feature extractor, with the algorithm \(a^*\) being the predicted best solver.
  
AAS largely aims to specify and apply the most suitable algorithm from a portfolio for solving a given problem instance. 
% Its standard rationale comes from the \textit{No Free Lunch} theorem, which asserts that no algorithm is universally superior on all possible instances~\cite{585893}.   
Typical AAS pipelines extract instance descriptors and combine them with empirical performance data, such as runtime or solution quality, to train predictive models that map instances to algorithms~\cite{alissa2023automated}, as depicted in Figure~\ref{AS_process}.

% Beyond traditional feature-based approaches, convolutional neural networks (CNNs) have been successfully applied to AAS~\cite{loreggia2016deep}. By learning from visual or structured representations of problem instances, CNNs eliminate the need for manual feature engineering, offering a more flexible and potentially more efficient alternative.

Beyond handcrafted numerical descriptors, representation learning has been explored for AAS. For example, applying CNNs to image-like encodings of combinatorial problem instances, constructed from instance descriptions, enables selection without manual feature engineering \cite{loreggia2016deep}. 
This paradigm is \emph{feature-free} in the sense of avoiding explicit engineered descriptors, but it does not rely on black-box probing nor on visualizing fitness landscapes concerning solution spaces.

% AAS has widespread applications across both discrete and continuous optimization. For discrete problems, the Traveling Salesman Problem (TSP) serves as a well-known benchmark~\cite{HUERTA2022115948}. In continuous optimization, AAS has been commonly applied to single-objective black-box optimization problems (BBO) such as those in the BBOB test suite.

% A widely used methodology in continuous optimization combines \textbf{Exploratory Landscape Analysis (ELA)} with machine learning models to predict the most suitable solver for a given problem instance~\cite{kerschke2019automated1}. ELA extracts low-level landscape features, such as convexity, curvature, and levelset, from a small sample of the problem, capturing important aspects like multimodality and separability~\cite{seiler2024learned}. These features are then used to construct AAS models.

% More recently, \textbf{Deep-ELA} has been introduced as a hybrid approach to improve upon traditional ELA by integrating deep learning techniques~\cite{seiler2025deep}. Deep-ELA employs pre-trained Transformer models to generate low-correlation, high-signal-to-noise-ratio (SNR) features, learned from millions of randomly generated optimization problems via self-supervised learning. This enables the method to capture problems' internal characteristics, replacing handcrafted ELA features with deep, automatically learned representations. Deep-ELA has proven to be a significant advancement, offering more robust and flexible AAS by leveraging the power of deep learning for landscape representation.

In continuous BBO, AAS has been widely studied under benchmarked settings such as BBOB~\cite{bbob2019}. 
A dominant approach is \textbf{Exploratory Landscape Analysis (ELA)}, which computes numerical landscape features such as curvature, convexity, and level-set structure, from a limited sample of evaluations and uses these features to train AAS models~\cite{kerschke2019automated1,seiler2024learned}. 
More recently, \textbf{Deep-ELA} replaces handcrafted ELA vectors with embeddings learned by pretrained transformers, trained via self-supervision on a large portfolio of synthetic optimization problems, yielding lower-correlation and higher-signal-to-noise-ratio representations for downstream selection~\cite{seiler2025deep}. 
Within this line, the central design choice is the representation of an instance under a probing budget.
% Classical ELA emphasizes explicitly computed statistics, whereas Deep-ELA focuses on learned numerical embeddings. 
The present work follows the same AAS setting but considers a complementary representation that preserves spatial structure by mapping probed evaluations to contour-plot images, enabling CNN-based selection from visual landscape information. 

%%%%%%%%%%%%%%%%%%%%%%%%%%%%%%%%%%%%%%%%%%
%%%%%%%%%%%%%%%%%%%%%%%%%%%%%%%%%%%%%%%%%%
\section{Methodology}
\label{sec:Methods}
%%%%%%%%%%%%%%%%%%%%%%%%%%%%%%%%%%%%%%%%%%
%%%%%%%%%%%%%%%%%%%%%%%%%%%%%%%%%%%%%%%%%%

\subsection{Benchmarks and Performance Metrics}
This study uses two benchmarks, respectively for single-objective optimization (SOO) and multi-objective optimization (MOO), both mirroring the evaluation settings of Deep-ELA~\cite{seiler2025deep}.

\textbf{Single-objective optimization.}
The primary evaluation dataset is the widely used COCO/BBOB 2009 noiseless single-objective benchmark suite~\cite{bbob2019}.
It consists of 24 scalable continuous functions grouped by properties such as separability, conditioning, and multimodality.
It additionally supports multiple instances via random shifts in both decision space and objective space.
% The experimental setup follows~\cite{seiler2025deep}, building on the benchmark protocol of~\cite{kerschke2019automated1}.
Following~\cite{seiler2025deep} (originally built on~\cite{kerschke2019automated1}), a fixed portfolio of 12 representative and complementary optimizers is adopted from COCO~\cite{hansen2021coco}: deterministic baselines, multi-level/model-based solvers,
CMA-ES variants, and OQNLP, enumerated in \cite{kerschke2019automated1}.
% deterministic baselines (BSrr, BSqi), multi-level/model-based solvers (MLSL, fmincon, HMLSL, MCS, SMAC-BBOB), CMA-ES variants (CMA-CSA, IPOP400D, HCMA), and OQNLP~\cite{povsik2015dimension, pal2013benchmarking, huyer2009benchmarking, pal2013comparison, atamna2015benchmarking, auger2013benchmarking, loshchilov2013bi, hutter2013evaluation}.
The raw benchmark comprises 480 instances (24 functions $\times$ 4 dimensions in \{2,3,5,10\} $\times$ 5 instances).
A \emph{problem configuration} refers to a function-dimension pair \((f,d)\), yielding 96 configurations, each associated with five instance-specific views.

Performance is measured by the Expected Running Time ($\mathrm{ERT}$) to a fixed target precision \(\varepsilon=10^{-2}\):
\[
\mathrm{ERT}(\varepsilon) = \frac{\sum_{i} \mathrm{FE}_i(\varepsilon)}{\sum_{i} \mathrm{Success}_i(\varepsilon)},
\]
where \(\mathrm{FE}_i(\varepsilon)\) is the number of evaluations used on instance \(i\) to reach the target, and \(\mathrm{Success}_i(\varepsilon)\in\{0,1\}\) is the binary success indicator.
To compare across heterogeneous difficulties, relative $\mathrm{ERT}$ ($\mathrm{relERT}$) is used. 
It normalizes $\mathrm{ERT}$ over the algorithm portfolio $\mathcal{A}$ per $(f, d)$ pair by the minimal $\mathrm{ERT}$:
\[
\mathrm{relERT}_{f,d,a} = \frac{\mathrm{ERT}_{f,d,a}}{\min\limits_{a' \in \mathcal{A}} \mathrm{ERT}_{f,d,a'}}.
\]
$\mathrm{ERT}$ is aggregated across the five instances per \((f,d)\) before computing $\mathrm{relERT}$, giving 96 $\mathrm{relERT}$ values per algorithm.
Undefined values due to missing/failed runs are imputed using a PAR10-style penalty (10$\times$ the maximum finite $\mathrm{relERT}$ observed, 36,690.3)~\cite{kerschke2019automated1}, to preserve comparability while penalizing non-performing algorithms.

\textbf{Multi-objective optimization.}
Downstream evaluation additionally considers bi-objective problems following the Deep-ELA evaluation setting~\cite{seiler2025deep} (itself akin to \cite{rook2022potential}).
The problem set contains 33 bi-objective instances with two-dimensional decision and objective spaces (\(d=2,m=2\) for objective dimension $m$):
ZDT~\cite{zitzler2000comparison}, DTLZ~\cite{deb2005scalable}, and MMF~\cite{yue2019novel} instances (excluding ZDT5 and MMF13), plus Bi-BBOB~\cite{brockhoff2022using} instances $f_{46}$, $f_{47}$, and $f_{50}$.
The algorithm portfolio follows that setting: NSGA-II~\cite{deb2002fast}, SMS-EMOA~\cite{beume2007sms}, MOEA/D~\cite{zhang2007moea}, and four additional optimizers (Omni-Optimizer~\cite{deb2005omni}, MOLE~\cite{schapermeier2022r}, MOGSA~\cite{grimme2019multimodality}, HIGA-MO~\cite{wang2017hypervolume}).

Performance is measured by the hypervolume ($\mathrm{HV}$) attained within a budget of 20{,}000 function evaluations.
For most instances, the $\mathrm{HV}$ reference point is pre-specified; when unavailable, it is derived per instance from the least favorable solution among all algorithms.
To compare across instances, $\mathrm{HV}$ is normalized using the best $\mathrm{HV}$ identified from additional runs of all algorithms with a budget of 100{,}000 evaluations.
Following Deep-ELA, normalized $\mathrm{HV}$ is then converted into relative $\mathrm{HV}$ by contrasting against the VBS-SBS gap:
\[
\mathrm{relHV}_{i, a}=\frac{\mathrm{HV}_{i, a}-\mathrm{HV}_{\text{SBS}}+\epsilon}{\mathrm{HV}_{\text{VBS}}-\mathrm{HV}_{\text{SBS}}+\epsilon},
\]
with \(\epsilon=10^{-8}\).
Values near 0 correspond to SBS-level performance, values near 1 correspond to VBS-level performance, and negative values indicate performance worse than SBS.

%%%%%%%%%%%%%%%%%%%%%%%%%%%%%%%%%%%%%%%%%%%%%%
\subsection{Visual Representations via Contour Plots}
%%%%%%%%%%%%%%%%%%%%%%%%%%%%%%%%%%%%%%%%%%%%%%

\textbf{SOO contour plots.}
Each SOO instance is represented as a 2D grayscale contour plot generated by evaluating the objective on a fixed uniform grid of resolution \(300\times 300\).
For 2D problems, this grid is naturally defined on the ambient $[-5,5]^2$ space.
For higher-dimensional problems (\(d\in\{3,5,10\}\)), a 2D slice-based representation is used.
Two coordinates are selected uniformly at random to span a 2D cross-sectional subspace $[-5,5]^2\subset[-5, 5]^D$ where the grid is situated, while all remaining coordinates are fixed to zero to leverage BBOB’s symmetric domain.
% The slice coordinate pair is sampled once per instance with a fixed seed and kept fixed throughout training and evaluation.
Each grid is evaluated using the official COCO implementation, and the resulting scalar field is normalized to $[0, 1]$ per plot and rendered as a filled contour plot.
This yields a consistent 2D representation while preserving a partial view of the full \(d\)-dimensional landscape and informative geometric structures.

To study the effect of input fidelity on learning, alongside the original input resolution $r=300$, generated plots are resized by interpolation to coarser CNN inputs with \(r\in\{64,128\}\) for additional experiments. 
Each SOO configuration \((f,d)\) provides five plots (one per instance), which are concatenated to an input tensor $X\in\mathbb R^{5\times r\times r}$ for that configuration. 

\begin{figure}[t]
   \centering
   \includegraphics[width=\linewidth]{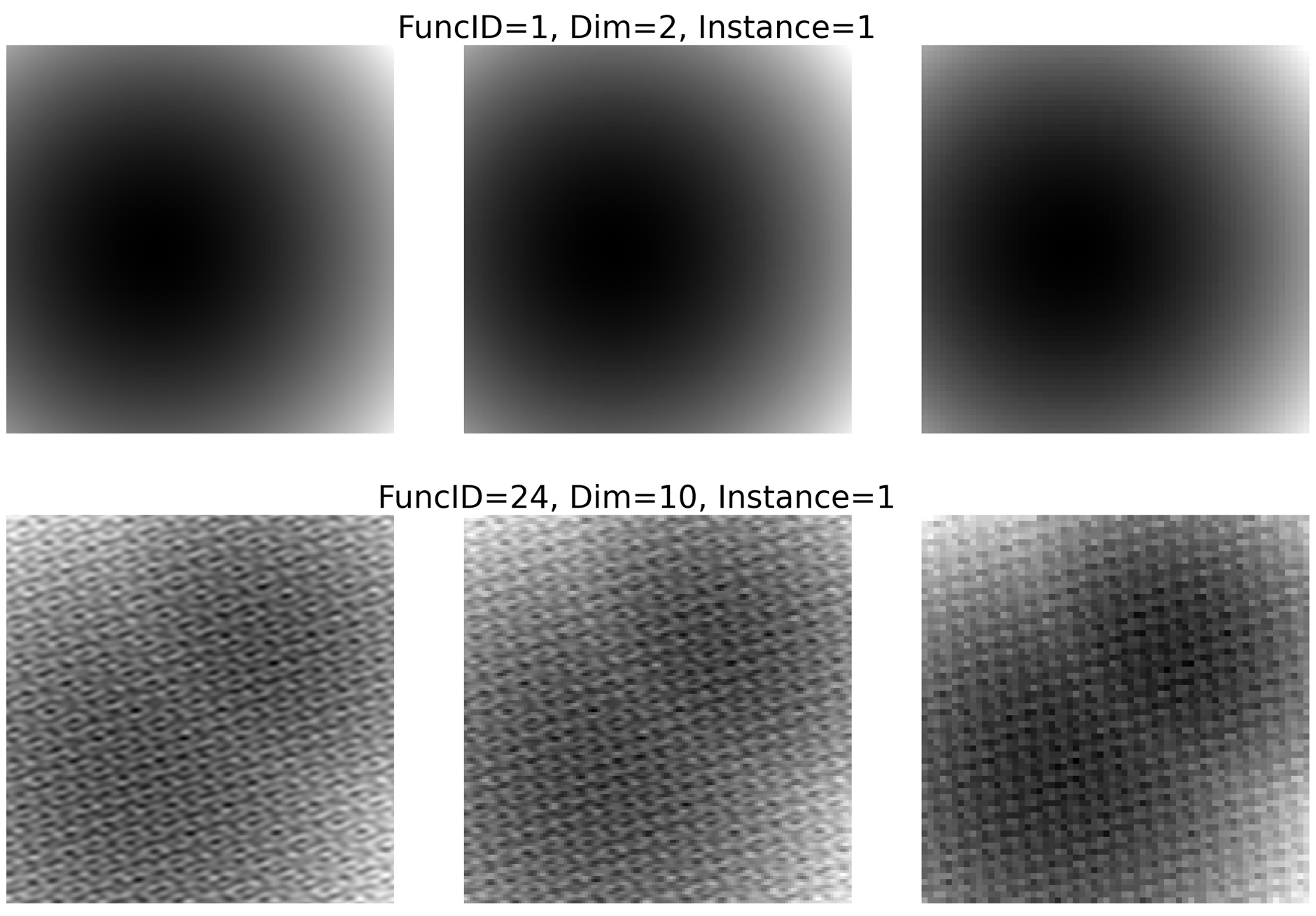}
   \caption{Contour plots for different problems. From left to right, the \emph{input} resolution decreases: 300$\times$300, 128$\times$128, and 64$\times$64 pixels.}
   \label{contour_com}
\end{figure}

The contouring step defines the probing cost: each contour plot is generated by evaluating the objective on a fixed \(300\times 300\) grid (i.e., \(90{,}000\) evaluations per plot), resulting in \(450{,}000\) evaluations per configuration.
% this resizing does not change the probing budget.
Figure~\ref{contour_com} illustrates examples at different resized input resolutions.
While the probing budget is large in evaluation-count terms, the measured preprocessing overhead on the BBOB benchmark implementation is modest in elapsed time, as Table~\ref{tab:contour_time} shows sub-second average generation times per plot. This makes the current design acceptable for offline proof-of-concept analysis, even though it is not intended for strict low-budget online deployment.
%Although the number of evaluations per configuration appears large, the practical cost is explicitly quantified via generation time and reports an average cost of under 0.3 seconds
%(Table~\ref{tab:contour_time}).
% (Appendix~\ref{app:generation_time}).

\begin{table}[t]
\centering
\caption{Average wall-clock time (seconds) to generate a single $300\times 300$ contour plot, by function group and dimension.}
\begin{tabular}{|l|c|c|}
\hline
\textbf{Function Group} & \textbf{Dimension} & \textbf{Avg. time per plot (s)} \\
\hline
F1--F5   & 2D & 0.18 \\ & 3D & 0.21 \\ & 5D & 0.21 \\ & 10D & 0.27 \\
\hline
F6--F9   & 2D & 0.17 \\ & 3D & 0.19 \\ & 5D & 0.19 \\ & 10D & 0.22 \\
\hline
F10--F14 & 2D & 0.17 \\ & 3D & 0.19 \\ & 5D & 0.21 \\ & 10D & 0.25 \\
\hline
F15--F19 & 2D & 0.20 \\ & 3D & 0.24 \\ & 5D & 0.29 \\ & 10D & 0.40 \\
\hline
F20--F24 & 2D & 0.31 \\ & 3D & 0.28 \\ & 5D & 0.29 \\ & 10D & 0.47 \\
\hline
\end{tabular}
\label{tab:contour_time}
\end{table}

\textbf{MOO sub-contour sampling.}
In the bi-objective setting considered here, the decision space is two-dimensional, so each objective can be visualized without slicing.
Since each MOO configuration represents one problem instance, to provide multiple informative views per instance and to reduce reliance on a single global rendering, a sub-contour-plot sampling strategy is adopted.
For each instance, five axis-aligned rectangular regions are sampled uniformly within the domain; each region is a window proportionally shrunk by a scaling coefficient \(\lambda_{\text{MOO}}\) from the instance's domain (with empirically chosen $\lambda_{\text{MOO}}=0.1$).
Similar to SOO, within each region, two objectives are evaluated on a \(300\times 300\) grid, interpolated to obtain additional plots with $r\in\{64/128\}$, and normalized per plot, yielding two contour-plot views over the same window. 
With five maps per objective, each input consists of two multi-channel tensors $X_{\text{obj1}}, X_{\text{obj2}}\in\mathbb R^{5\times r\times r}$.
This sampling process is repeated 20 times per instance, consistent with the evaluation design of Deep-ELA~\cite{seiler2025deep}.

\begin{figure}[t]
    \centering
    \begin{minipage}[t]{\linewidth}
        \centering
        \includegraphics[width=\linewidth]{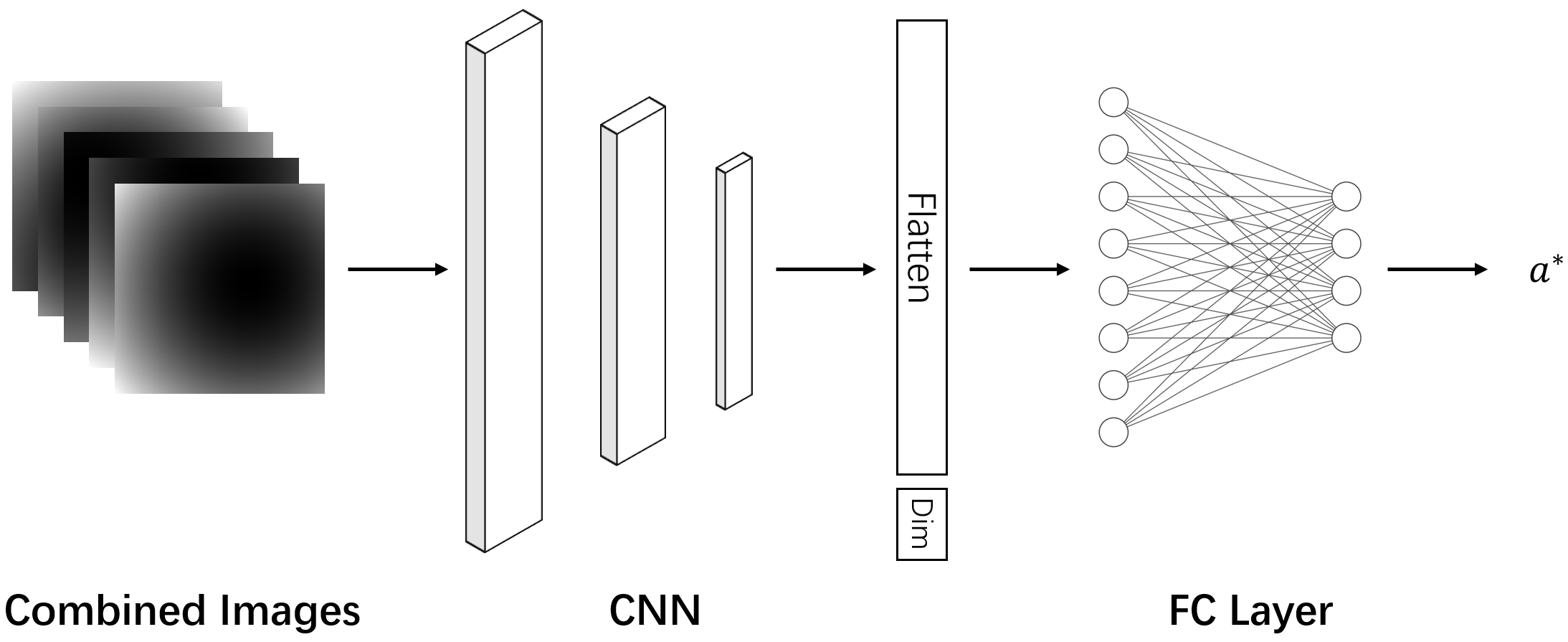}
        \caption{Illustration of the combined model. The stacked contour tensor is encoded into feature vector $z$, which is passed to the regression head to predict per-algorithm performance and select the best solver $a^{*}$}
        \label{combined_model}
    \end{minipage}
    \hfill
    \begin{minipage}[t]{\linewidth}
        \centering
        \includegraphics[width=\linewidth]{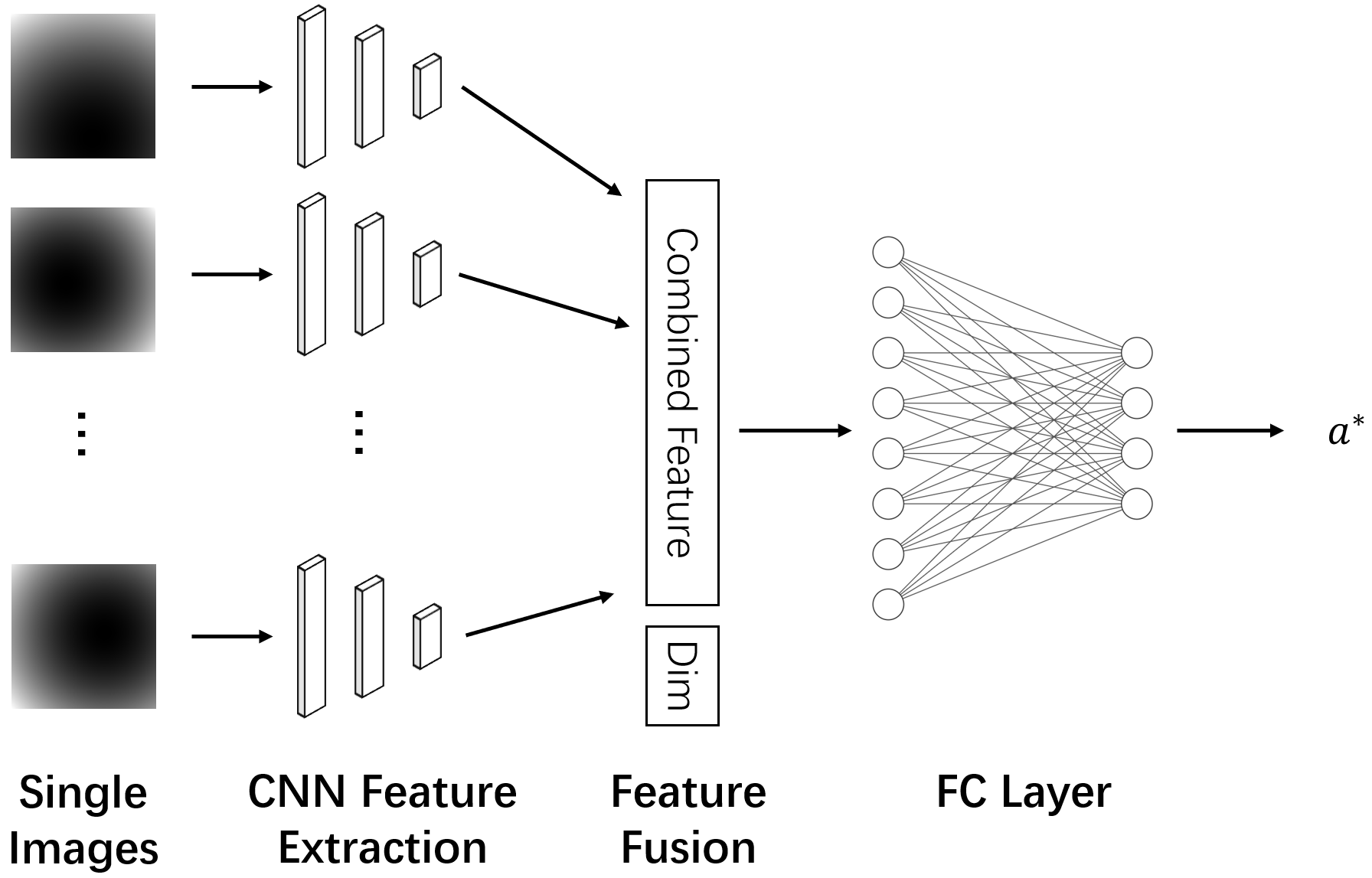}
        \caption{Illustration of the separate model. Each contour view is encoded separately into a feature vector, and the resulting features are concatenated into $z$, which is passed to the regression head to predict per-algorithm performance and select the best solver $a^{*}$}
        \label{separate_model}
    \end{minipage}
\end{figure}

%%%%%%%%%%%%%%%%%%%%%%%%%%%%%%%%%%%%%%%%%%%%%%
\subsection{CNN Architectures for Algorithm Selection}
%%%%%%%%%%%%%%%%%%%%%%%%%%%%%%%%%%%%%%%%%%%%%%
The architecture consists of a vision encoder and a regression head that predicts per-algorithm performance.
Given 5 views \(X\in\mathbb R^{5\times r\times r}\), two variants are considered:
\begin{enumerate}
    \item \emph{Combined model}: The five views are stacked along the channel dimension, with random view ordering to avoid spurious dependence on stacking order. The encoder operates on the stacked view directly to extract the feature vector $z$. (Figure~\ref{combined_model})
    \item \emph{Separate model}: Each view $x_i\in\mathbb R^{r\times r}$ is individually processed by a shared encoder, and the resulting features are concatenated to form $z$. (Figure~\ref{separate_model})
\end{enumerate}
For SOO, the encoder is a three-layer CNN; the resulting feature vector is appended with the dimension \(d\) and fed to a regression head predicting per-solver \(\mathrm{relERT}\).
For MOO, ResNet-18 is used as the encoder due to the higher representational demands observed empirically. 
The same encoder is applied to \(X_{\text{obj1}}\) and \(X_{\text{obj2}}\) with shared weights, producing embeddings \(z_{\text{obj1}}, z_{\text{obj2}}\) that are concatenated and passed to an analogous regression head predicting per-solver \(\mathrm{relHV}\).

Input resolutions \(r \in \{64,128,300\}\) are evaluated for both models on both SOO and MOO benchmarks. The only exception is the separate model at \(r = 300\) under SOO evaluation, which is omitted due to training-cost considerations.

The overall procedure is summarized in Algorithm~\ref{algo:cnn_combined_separate} and follows the same selection principle, i.e., predict performance over the portfolio and select by \(\arg\min_a\mathrm{relERT}_a\) or \(\arg\max_a\mathrm{relHV}_a\).

\begin{algorithm}[t]
\caption{CNN-Based Algorithm Selection (Combined vs.\ Separate)}
\label{algo:cnn_combined_separate}
\begin{algorithmic}[1]
\Statex \textbf{Input:} Contour-plot tensor $X\in\mathbb R^{k\times r\times r}$ (SOO), dimension $d$, and $k{=}5$; portfolio $\mathcal A{=}\{a_1,\dots,a_m\}$.
\Statex \textbf{Output:} Selected algorithm $a^*$.
\Statex \textbf{Combined model:}
\State $z \leftarrow \Phi_{\text{cnn}}(X)$ 
\State $\hat{\mathcal P} \leftarrow \Psi(z, d)$ \Comment{$\hat{\mathcal P}\in\mathbb R^{m}$}
\State $a^* \leftarrow \arg\min_{\alpha\in\{1,\dots,m\}} \hat{\mathcal P}_\alpha$ \Comment{SOO (relERT); use $\arg\max$ for MOO (relHV)}
\Statex \textbf{Separate model:}
\State $z_j \leftarrow \Phi_{\text{cnn}}(x^{(j)})$ for $j=1,\dots,k$ \Comment{$\Phi$ shared}
\State $z \leftarrow \text{Concat}(z_1,\dots,z_k)$
\State $\hat{\mathcal P} \leftarrow \Psi(z, d)$ \Comment{$\hat{\mathcal P}\in\mathbb R^{m}$}
\State $a^* \leftarrow \arg\min_{\alpha\in\{1,\dots,m\}} \hat{\mathcal P}_\alpha$ \Comment{SOO; $\arg\max$ for MOO}
\Statex \textbf{MOO note:} if using two objectives, replace $X$ by $(X^{(1)},X^{(2)})$; encode each to $z^{(1)},z^{(2)}$ and concatenate before $\Psi$.
\end{algorithmic}
\end{algorithm}

%%%%%%%%%%%%%%%%%%%%%%%%%%%%%%%%%%%%%%%%%%
%%%%%%%%%%%%%%%%%%%%%%%%%%%%%%%%%%%%%%%%%%
\section{Computational Analysis}
\label{sec:Experiments}
%%%%%%%%%%%%%%%%%%%%%%%%%%%%%%%%%%%%%%%%%%
%%%%%%%%%%%%%%%%%%%%%%%%%%%%%%%%%%%%%%%%%%

\begin{table*}[t]
\centering
\caption{$\mathrm{relERT}$ values of different models. SBS is the HCMA algorithm,  The Deep-ELA Medium Model cannot handle data more than 6 dimensions, resulting in empty space for D=10. The last five columns are the results of our CNN-based AAS models.}
\label{res_table}    
\begin{tabular}{cc|c|cc|cc|cc||ccc|ccc}
\\
\toprule
\textbf{D} & \textbf{F. Group} & \textbf{SBS} & \multicolumn{2}{c|}{\textbf{ELA (50d)*}} & \multicolumn{2}{c|}{\textbf{Large (25d)}*} & \multicolumn{2}{c||}{\textbf{Medium (25d)*}} & \multicolumn{3}{c|}{\textbf{Combined CNN}} & \multicolumn{2}{c}{\textbf{Separate CNN}}\\
 &  &  & RF & MLP & kNN & RF & kNN & RF & 64×64 & 128×128 & 300×300 & 64×64 & 128×128\\
\midrule
\multirow{6}{*}{2}
& 1 & \textbf{3.71} & 10.41 & 10.59 & 9.24 & 10.30 & 17.57 & 7.57 & 14.65 & 14.65 & 4.12 & 14.89 & 14.65\\
& 2 & 5.80 & 8.51 & 3.72 & 2.65 & 2.87 & 2.69 & 3.37 & \textbf{1.66} & 1.74 & 11.52 & \textbf{1.66} & \textbf{1.66}\\
& 3 & 6.29 & 1473.16 & 4.72 & 3.52 & 3.12 & 3.91 & 4.40 & 1.27 & \textbf{1.0} & \textbf{1.0} & \textbf{1.0} & \textbf{1.0}\\
& 4 & 25.34 & \textbf{3.89} & 9.25 & 6.45 & 9.76 & 5.73 & 5.99 & 8.08 & 6.86 & 6.20 & 8.08 & 8.08\\
& 5 & 44.95 & 148.39 & \textbf{3.32} & 3.69 & 5.06 & 3.79 & 8.08 & 3.61 & 5.60 & 4.09 & 3.87 & 3.87\\
\cmidrule{2-14}
& all & 17.69 & 342.22 & 6.43 & 5.21 & 6.36 & 6.91 & 5.99 & 6.03 & 6.15 & \textbf{5.13} & 6.08 & 6.03\\
\midrule

\multirow{6}{*}{3}
& 1 & 356.10 & 1480.68 & 11.87 & 19.94 & 66.76 & 11.72 & 53.23 & \textbf{4.80} & 5.54 & 6.22 & 5.54 & 6.84\\
& 2 & 4.46 & 8.33 & 3.50 & 2.73 & 3.02 & 2.67 & 2.87 & 5.32 & 1.90 & 3.50 & \textbf{1.83} & \textbf{1.83}\\
& 3 & 4.98 & 7.07 & 3.82 & 2.69 & 3.48 & 2.59 & 3.53 & 3.06 & \textbf{2.29} & 2.92 & 4.98 & 2.85\\
& 4 & \textbf{2.63} & 441.96 & 5.06 & 5.15 & 4.63 & 5.36 & 4.83 & 8.90 & 5.21 & 3.95 & \textbf{2.63} & 8.90\\
& 5 & 66.81 & \textbf{1.22} & 2.54 & 30.75 & 12.02 & 2.70 & 4.99 & 66.30 & 51.57 & 2.83 & 51.39 & 2.73\\
\cmidrule{2-14}
& all & 90.43 & 403.67 & 5.44 & 12.65 & 18.61 & 5.10 & 14.35 & 18.19 & 13.78 & \textbf{3.90} & 13.75 & 4.75\\
\midrule

\multirow{6}{*}{5}
& 1 & 11.99 & 14.14 & \textbf{11.97} & 17.01 & 17.31 & 17.50 & 17.85 & 16.47 & 11.99 & 11.99 & 11.99 & 11.99\\
& 2 & 3.90 & 369.26 & 2.62 & 2.40 & 4.27 & 2.48 & 2.45 & 5.06 & \textbf{1.76} & 3.90 & 3.90 & 3.90\\
& 3 & 4.21 & 150.44 & 3.97 & 3.70 & 4.87 & 3.76 & 3.69 & 3.31 & \textbf{3.07} & 4.21 & 4.21 & 4.21\\
& 4 & 4.29 & 1470.28 & 6.81 & \textbf{1.87} & 4.07 & 3.95 & 3.51 & 4.29 & 3.68 & 4.29 & 4.29 & 4.29\\
& 5 & 7.67 & 1.13 & 1.83 & \textbf{1.08} & 4.73 & 1.72 & 1.43 & 2.99 & 7.67 & 7.67 & 7.67 & 7.67\\
\cmidrule{2-14}
& all & 6.52 & 402.38 & 5.56 & \textbf{5.47} & 7.17 & 6.02 & 5.93 & 6.48 & 5.80 & 6.52 & 6.52 & 6.52\\
\midrule

\multirow{6}{*}{10}
& 1 & \textbf{2.74} & 14.64 & 15.27 & 9.54 & 9.53 & - & - & \textbf{2.74} & 3.44 & \textbf{2.74} & \textbf{2.74} & \textbf{2.74}\\
& 2 & 2.16 & \textbf{1.62} & 1.76 & 2.40 & 2.43 & - & - & 2.16 & 2.16 & 2.16 & 2.16 & 2.16\\
& 3 & \textbf{2.76} & 2.87 & 4.35 & 3.54 & 3.62 & - & - & \textbf{2.76} & \textbf{2.76} & \textbf{2.76} & \textbf{2.76} & \textbf{2.76}\\
& 4 & 2.02 & 442.01 & \textbf{1.96} & 2.06 & 2.05 & - & - & 2.02 & 2.02 & 2.02 & 2.02 & 2.02\\
& 5 & 23.64 & 148.01 & 3.25 & \textbf{1.73} & 11.33 & - & - & 23.64 & 23.64 & 23.64 & 23.64 & 23.64\\
\cmidrule{2-14}
& all & 6.85 & 126.84 & 5.46 & \textbf{3.91} & 5.93 & - & - & 6.85 & 7.00 & 6.85 & 6.85 & 6.85\\
\midrule

\multirow{6}{*}{all}
& 1 & 93.63 & 379.97 & 12.43 & 14.10 & 25.97 & 15.60 & 26.22 & 9.67 & 8.90 & \textbf{6.27} & 8.79 & 9.05\\
& 2 & 4.08 & 96.93 & 2.90 & 2.54 & 3.15 & 2.62 & 2.90 & 3.55 & \textbf{1.89} & 5.27 & 2.39 & 2.39\\
& 3 & 4.56 & 408.38 & 4.21 & 3.36 & 3.77 & 3.42 & 3.88 & 2.60 & \textbf{2.28} & 2.72 & 3.24 & 2.71\\
& 4 & 8.57 & 589.54 & 5.77 & 3.88 & 5.13 & 5.01 & 4.78 & 5.82 & 4.44 & \textbf{4.12} & 4.25 & 5.82\\
& 5 & 35.77 & 74.69 & \textbf{2.74} & 9.31 & 8.29 & \textbf{2.73} & 4.83 & 24.14 & 22.12 & 9.56 & 21.64 & 9.48\\
\cmidrule{2-14}
& all & 30.37 & 318.78 & 5.72 & 6.81 & 9.52 & 6.01 & 8.76 & 9.39 & 8.18 & \textbf{5.60} & 8.30 & 6.04\\
\bottomrule
\end{tabular}

\vspace{2pt}
\parbox{\linewidth}{\footnotesize * The results of ELA-based models, ELA (50d), are taken from~\cite{prager2022automated} and those of Deep-ELA based models, Large (25d) and Medium (25d), are taken from~\cite{seiler2025deep}.}
\end{table*}

% %%%%%%%%%%%%%%%%%%%%%%%%%%%%%%%%%%%%%%%%%%%%%%
\subsection{Experiment Setting}
% %%%%%%%%%%%%%%%%%%%%%%%%%%%%%%%%%%%%%%%%%%%%%%

\textbf{SOO evaluation.}
Following the evaluation setting of \cite{kerschke2019automated1}, leave-one-out cross-validation is performed over the 96 SOO configurations: each fold holds out one \((f,d)\) configuration and trains on the remaining 95.

\textbf{MOO evaluation.}
Consistent with the evaluation design of Deep-ELA~\cite{seiler2025deep}, for each bi-objective instance, 20 repetitions of the window-sampling procedure are generated, where 15 repetitions are used for training and the remaining 5 for testing.
% This protocol evaluates robustness of selection to stochastic window sampling on the same underlying instance, rather than targeting an out-of-distribution generalization study across unseen instances.

\textbf{Implementation detail.}
Model training uses the same implementation across folds, with hyper-parameters selected via a lightweight grid search on a validation split within each fold.
All models are trained on a workstation equipped with 2 AMD EPYC 7763 64-Core CPUs, 8 NVIDIA RTX 3090 GPUs, and 256 Gigabytes of RAM, using PyTorch for implementation. 
Due to the modest model size, each model is trained on a single GPU.

%%%%%%%%%%%%%%%%%%%%%%%%%%%%%%%%%%%%%%%%%%%%%%
\subsection{Baselines}
%%%%%%%%%%%%%%%%%%%%%%%%%%%%%%%%%%%%%%%%%%%%%%

Performance is compared against the single best solver (SBS), defined as the portfolio algorithm with the lowest mean $\mathrm{relERT}$ over the 96 configurations, representing the best static choice. 
In the current setup, SBS corresponds to HCMA.
Feature-based AAS baselines include two ELA-based selectors, with multilayer-perceptron (MLP) and random forest (RF) heads respectively, from \cite{prager2022automated} and Deep-ELA variants, with RF and kNN heads, from \cite{seiler2025deep}. For SOO, the $25d$ Deep-ELA variants are reported, as they exhibit more consistent performance across function groups and dimensions than their $50d$ counterparts, while for MOO all variants are included. 
Deep-ELA Medium is constrained by its combined dimensionality limit \(d{+}m\le 6\), and therefore SOO results for \(d{=}10\) (\(m{=}1\)) are outside of its scope.

%%%%%%%%%%%%%%%%%%%%%%%%%%%%%%%%%%%%%%%%%%
\subsection{Single-objective optimization}
%%%%%%%%%%%%%%%%%%%%%%%%%%%%%%%%%%%%%%%%%%

\begin{figure*}[t]
   \centering  
   \includegraphics[width=0.65\linewidth]{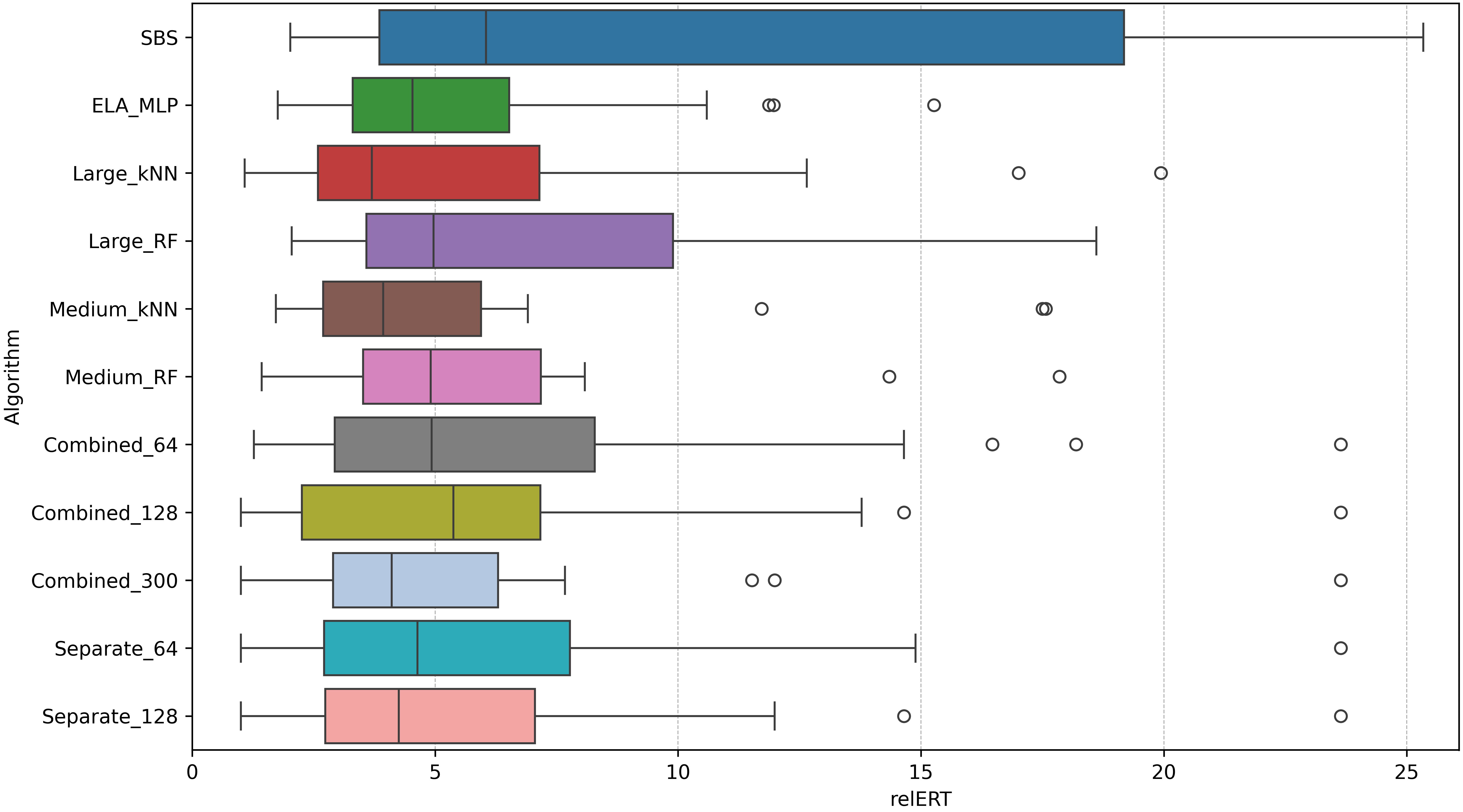} 
   \caption{$\mathrm{relERT}$ distribution per algorithm across all function groups and dimensions.}
   \label{relert_boxplot}
\end{figure*}

Table~\ref{res_table} reports $\mathrm{relERT}$ aggregated by dimension and function group.
Across all configurations, the combined CNN improves substantially over SBS (mean $\mathrm{relERT}$ \(30.37\rightarrow 5.60\) at \(r{=}300\)), while remaining competitive with the strongest feature-based baselines (ELA-MLP: \(5.72\)).
The separate CNN also improves over SBS (e.g., \(6.04\) at \(r{=}128\)), though its gains are less consistent across groups.

Performance varies by dimension and function group.
The most pronounced gains occur at \(d{=}3\), where the combined CNN at \(r{=}300\) attains mean $\mathrm{relERT}$ \(3.90\), outperforming the reported ELA and Deep-ELA baselines in the same slice of the table.
Across function groups, CNN-based selectors perform strongly on groups 1--4 (F1--F19), while performance on group 5 (F20--F24) remains mixed, where some feature-based methods retain an advantage.

Input fidelity matters: for the combined model, increasing \(r\) typically reduces mean $\mathrm{relERT}$ (e.g., overall \(9.39\rightarrow 8.18\rightarrow 5.60\) for \(r{=}64,128,300\)).
This trend is consistent with the boxplot in Figure~\ref{relert_boxplot}, where the combined CNN at \(r{=}300\) exhibits a lower median and a tighter distribution than its lower-\(r\) counterparts.
These gains come with increased training cost; hence \(r\) represents a practical accuracy--cost trade-off rather than a change in probing budget.

To assess robustness across aggregated settings, a paired Wilcoxon signed-rank test is conducted over the 20 aggregated $\mathrm{relERT}$ entries (4 dimensions \(\times\) 5 function groups) comparing the combined CNN (\(r{=}300\)) against selected baselines.
The combined CNN significantly improves over SBS (\(p=0.037\)).
Against ELA-MLP and Deep-ELA Medium-kNN, there is no statistically significant difference over all 20 entries (\(p=0.286\) and \(p=0.598\), respectively), while restricting to groups 1--4 yields stronger evidence against ELA-MLP (\(p=0.05\)) and remains inconclusive against Deep-ELA Medium-kNN (\(p=0.339\)).
Given the small number of aggregated test points, these tests are best interpreted as supportive rather than definitive.

\begin{table*}[t]
\centering
\caption{$\mathrm{relHV}$ of function groups (rows) over different methods/settings (columns).}
\label{tab:moo_comparison}
\resizebox{\linewidth}{!}{%
\begin{tabular}{l|cc|cc|cc|cc||ccc|ccc}
\toprule
\textbf{Function group}
& \multicolumn{2}{c|}{\textbf{Large (25d)}}
& \multicolumn{2}{c|}{\textbf{Large (50d)}}
& \multicolumn{2}{c|}{\textbf{Medium (25d)}}
& \multicolumn{2}{c||}{\textbf{Medium (50d)}}
& \multicolumn{3}{c|}{\textbf{Combined CNN}}
& \multicolumn{3}{c}{\textbf{Separate CNN}} \\
% \cmidrule(lr){2-3}\cmidrule(lr){4-5}\cmidrule(lr){6-7}\cmidrule(lr){8-9}\cmidrule(lr){10-12}\cmidrule(lr){13-15}
& \textbf{kNN} & \textbf{RF}
& \textbf{kNN} & \textbf{RF}
& \textbf{kNN} & \textbf{RF}
& \textbf{kNN} & \textbf{RF}
& 64 & 128 & 300
& 64 & 128 & 300 \\
\midrule
\textbf{BiBBOB}
& \textbf{1.000} & \textbf{1.000}
& \textbf{1.000} & \textbf{1.000}
& \textbf{1.000} & \textbf{1.000}
& \textbf{1.000} & \textbf{1.000}
&  \textbf{1.000} & \textbf{1.000} & \textbf{1.000}
& \textbf{1.000} & \textbf{1.000} & \textbf{1.000} \\
\textbf{DTLZ}
& 0.914 & 0.457
& \textbf{1.000} & 0.651
& \textbf{1.000} & 0.651
& \textbf{1.000} & 0.822
& 0.819 & 0.914 & 0.876
& 0.924 & 0.924 & 0.981 \\
\textbf{MMF}
& 0.892 & 0.463
& 0.931 & 0.609
& 0.929 & 0.764
& \textbf{1.000} & 0.933
& 0.886 & 0.941 & 0.907
& 0.969 & 0.960 & 0.917 \\
\textbf{ZDT}
& -0.079 & -0.087
& 0.041 & 0.209
& 0.688 & 0.185
& 0.616 & 0.745
& 0.960 & \textbf{1.000} & \textbf{1.000}
& \textbf{1.000} & \textbf{1.000} & \textbf{1.000} \\
\midrule
\textbf{all}
& 0.682 & 0.458
& 0.743 & 0.617
& 0.904 & 0.650
& 0.904 & 0.875
& 0.933 & 0.964 & 0.946
& 0.973 & 0.971 & \textbf{0.974} \\
\bottomrule
\end{tabular}%
}
\end{table*}

\subsection{Multi-objective optimization}
Table~\ref{tab:moo_comparison} reports $\mathrm{relHV}$ on 33 bi-objective instances grouped by benchmark family, where values near 1 indicate closing the SBS--VBS gap and negative values indicate worse-than-SBS performance. Overall, both CNN variants are competitive and outperform the feature-based Deep-ELA baselines in the aggregate. The best mean $\mathrm{relHV}$ scores are achieved by the Separate CNN, ranging from $0.971$--$0.974$.
% , closely followed by \(r{=}64\) and \(r{=}128\) (\(0.973\) and \(0.971\)). 
For the Combined CNN, \(r{=}128\) performs best (\(0.964\)), with \(r{=}64\) and \(r{=}300\) slightly lower (\(0.933\) and \(0.946\)). Among the feature-based baselines, Deep-ELA Medium (kNN; 25d/50d) is strongest (\(0.904\)), while the remaining variants perform worse overall.

The breakdown by benchmark family highlights where gains arise. BiBBOB is saturated (all methods at \(1.0\)), leaving little discriminative room under this grouping. On ZDT, CNN-based selectors essentially solve the subset (Separate: \(1.0\) for all \(r\); Combined: \(0.960\) at \(r{=}64\) and \(1.0\) at \(r\ge128\)), whereas some Deep-ELA settings are near-zero or negative (e.g., Large-25d), indicating failures to outperform SBS. On DTLZ and MMF, CNN models remain strong but do not uniformly match the best Deep-ELA settings: Separate CNN reaches up to \(0.981\) on DTLZ (vs.\ \(1.0\) for Deep-ELA kNN, 50d) and up to \(0.969\) on MMF (vs.\ \(1.0\) for Deep-ELA Medium kNN, 50d).

Input resolution effects are modest in the overall averages but not monotone across families. The Separate CNN benefits from higher \(r\) on DTLZ (up to \(0.981\) at \(r{=}300\)) yet degrades on MMF at \(r{=}300\) (\(0.917\)), suggesting that finer-scale window detail can be informative for some landscapes but less stable under stochastic window sampling for others. Given the near-identical mean performance across \(r\), \(r{=}64\) is a reasonable practical default in MOO, with \(r\in\{128,300\}\) serving as sensitivity checks rather than a uniformly better setting.

%%%%%%%%%%%%%%%%%%%%%%%%%%%%%%%%%%%%%%%%%%
%%%%%%%%%%%%%%%%%%%%%%%%%%%%%%%%%%%%%%%%%%
\section{Discussion and Conclusion}
\label{sec:Conclusion}
%%%%%%%%%%%%%%%%%%%%%%%%%%%%%%%%%%%%%%%%%%
%%%%%%%%%%%%%%%%%%%%%%%%%%%%%%%%%%%%%%%%%%
This work studies automated algorithm selection for continuous black-box optimization using an image-based representation of problem instances. 
Each configuration is probed by generating contour maps from objective evaluations on a fixed $300\times 300$ grid (five instance-specific views per $(f,d)$), and a CNN-based regressor predicts per-solver performance for selection. 
Across the BBOB 2009 SOO benchmark under LOOCV, the best-performing variant (combined CNN with $r{=}300$ input) reduces mean relERT from 30.37 (SBS) to 5.60, and is comparable in aggregate to the strongest feature-based baseline reported in Table~\ref{res_table} (ELA-MLP: 5.72). 
Input resizing from the same probed field indicates that higher input fidelity tends to improve selection quality, at the cost of higher training-time and memory demands.

The same contour-based principle also extends to the bi-objective setting under the Deep-ELA protocol. Under this inherited evaluation setting, apart from BiBBOB which is saturated across all methods, CNN-based selectors remain competitive on relHV and perform particularly strongly on ZDT, while DTLZ and MMF remain closer to the strongest Deep-ELA variants. These results suggest that the image-based representation is not confined to the SOO setting. 
% Table~\ref{tab:moo_comparison} shows that CNN-based selectors are competitive on relHV, with the best overall score attained by the Separate CNN at \(r=300\) (0.974), closely followed by \(r=64\) (0.973).
% Performance is benchmark-dependent: ZDT is solved at relHV \(=1.0\) by the Separate CNN across input resolutions, while DTLZ and MMF remain close to, but do not uniformly exceed, the strongest Deep-ELA settings.
% BiBBOB is saturated (all methods at 1.0), suggesting limited discriminative room under this grouping. 

Several limitations bound the scope of the claims. 
First, the SOO probing budget is substantial (five $300\times 300$ maps per configuration), which is acceptable for offline dataset construction but mismatched to strict online probing regimes. This study focuses on representational feasibility rather than probing-budget optimality. Therefore, budget-aware probing and direct low-resolution evaluation are natural next steps.  
Second, for $d>2$, the representation is only a partial visual probe of the landscape. A single axis-aligned 2D slice with fixed coordinates may miss important high-dimensional structure, which plausibly contributes to the weaker performance on the hardest function group.
%the representation is a 2D slice with fixed coordinates, which may miss high-dimensional structure and plausibly contributes to weaker performance on the hardest function group. 
Third, both SOO and MOO evaluations are tied to a fixed portfolio and protocol (including window sampling for MOO), and the MOO study is limited to $d{=}2,m{=}2$. Future work should therefore prioritize cost-sensitive and multi-view probing designs, stronger high-dimensional representations such as multiple slices or hybrid visual--numerical features, and validation across alternative portfolios as well as optimization budgets.

\section*{Acknowledgment}

We thank Shiya Ye for assistance with the initial collection and organisation of the BBOB-related data used in this study.

\bibliographystyle{IEEEtran}
\bibliography{IJCNN2026/references}

% \clearpage
% \input{IJCNN2026/sections/appendix}

\end{document}